\documentclass{bmvc2k}
\usepackage{amssymb}
\usepackage{booktabs}
\pdfoutput=1

\title{Deep Perceptual Mapping for Thermal to Visible Face Recognition}

\addauthor{M. Saquib Sarfraz}{https://cvhci.anthropomatik.kit.edu/~ssarfraz}{1}
\addauthor{Rainer Stiefelhagen}{http://cvhci.anthropomatik.kit.edu}{2}

\addinstitution{
 Institute of Anthropomatics \& Robotics\\
 Karlsruhe institute of Technology\\
 Karlsruhe, Germany.
 }

\runninghead{Sarfraz, Stiefelhagen}{Deep Perceptual Mapping Thermal-Visible Face}

\def\eg{\emph{e.g}\bmvaOneDot}

\def\etal{\emph{et al}\bmvaOneDot}
\def\ie{\emph{i.e}\bmvaOneDot}
\def\wrt{\emph{w.r.t}\bmvaOneDot}
\begin{document}

\maketitle

\begin{abstract}
Cross modal face matching between the thermal and visible spectrum is a much desired capability for night-time surveillance and security applications. Due to a very large modality gap, thermal-to-visible face recognition is one of the most challenging face matching problem. In this paper, we present an approach to bridge this modality gap by a significant margin. Our approach captures the highly non-linear relationship between the two modalities by using a deep neural network. Our model attempts to learn a non-linear mapping from visible to thermal spectrum while preserving the identity information. We show substantive performance improvement on a difficult thermal-visible face dataset. The presented approach improves the state-of-the-art by more than 10\% in terms of Rank-1 identification and bridge the drop in performance due to the modality gap by more than 40\%. 
\end{abstract}
\section{Introduction}
\label{sec:intro}
Predominately, face recognition has been focused in the visible spectrum. This pertains to a large number of applications from biometrics, access control systems, social media tagging to person retrieval in multimedia. Among the main challenges in visible face recognition, the different lighting/illumination condition has proven to be one of the big factors for appearance change and performance degradation. Many prior studies \cite{li2007, socolinsky2002, nicolo2012, Klare13} have stated better face recognition performance in the infra-red spectrum because it is invariant to ambient lighting. Relatively recently, few efforts have been devoted in the cross-modal face recognition scenarios, where the objective is to identify a person captured in infra-red spectrum based on its stored high resolution visible face image. The motivation for this lies in the night-time or low light surveillance tasks where the image is captured discretely or covertly through active or passive infra-red sensors. In fact there does not exist a reliable solution for matching thermal faces, acquired covertly in such conditions, against the stored visible database such as police mugshots. This poses a significant research gap for such applications in aiding the law enforcement agencies. In the infra-red spectrum, thermal signatures emitted by skin tissues can be acquired through passive thermal sensors without using any active light source. This makes it an ideal candidate for covert night-time surveillance tasks.

As opposed to the visible spectrum wavelength (0.35$\mu m$ to 0.74$\mu m$), the infra-red spectrum lies in four main ranges. Near infra-red `NIR' (0.74$\mu m$-1$\mu m$), short-wave infra-red `SWIR' (1-3$\mu m$), mid-wave infra-red `MWIR' (3-5$\mu m$) and long-wave infra-red `LWIR' (8-14$\mu m$). Since the NIR and SWIR bands are reflection dominant, they are more close to the visible spectrum. The MWIR and LWIR are the thermal spectrum and are emission dominant i.e. dependent on material emissivity and temperature. Skin tissue has high emissivity in both the MWIR and LWIR spectrum. Because of this natural difference between reflective visible spectrum and sensed emissivity in the thermal spectrum, the images in the two modalities are very different and have a large modality gap. This hinders reliable face matching across the two domains. It is, perhaps, for this reason that most of the earlier studies, aiming at cross-modal face recognition, relies only in visible-to-NIR face matching. While achieving very good results, NIR imaging use an active light source that makes it redundant for covert night-time surveillance. More recently, some attempts have been made in thermal-to-visible face recognition, indicating a significant performance gap due to the very challenging nature of the problem and the large modality gap.

\begin{figure}
\centering
    \includegraphics[width=0.95\textwidth]{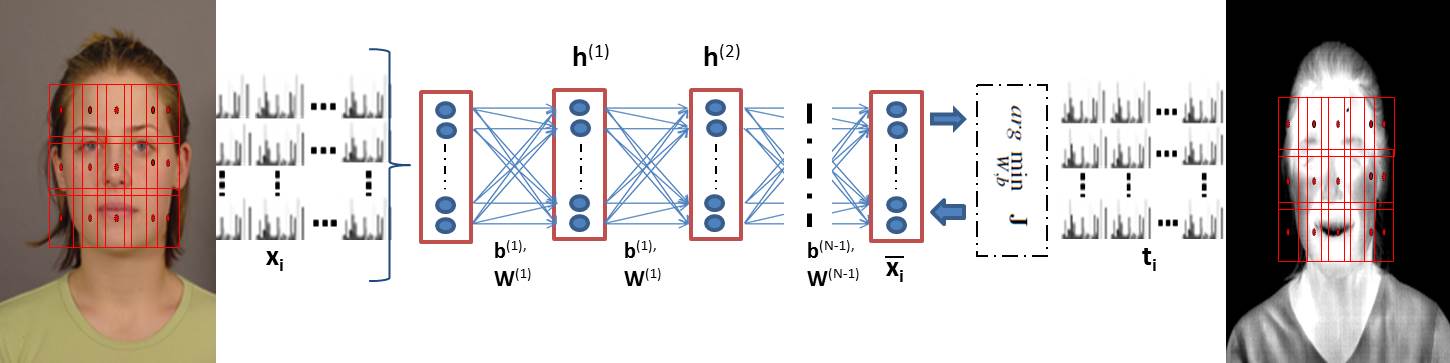}
    \caption{Deep Perceptual Mapping (DPM): densely computed features from the visible domain are mapped through the learned DPM network to the corresponding thermal domain.}
    \label{fig:1}
    \end{figure}

In this paper, we seek to bridge this gap by trying to model directly the highly non-linear mapping between the two modalities. Our contribution is a useful model, based on a feed-forward deep neural network, and its effective design steps in order to map the perceptual differences between the two modalities while preserving the identity information. We show that this mapping can be learned from relatively sparse data and works quite good in practice. Our model tries to learn a non-linear regression function between the visible and thermal data. The learned projection matrices capture the non-linear relationship well and are able to bring the two closer to each other. Another contribution is to further the best published state-of-the-art performance on a very challenging dataset (University of Notre Dame UND-X1) by more than 10\%. Our results show that this accounts for bridging the performance gap due to modality difference by more than 40\%. To the best of our knowledge, this is the first attempt in using deep neural networks to bridge the modality gap in thermal-visible face recognition. Figure \ref{fig:1} provides an overview of the approach.

We will start by discussing some of the related work in section \ref{related}. Section \ref{DPM} will detail the presented approach. We will conclude in section \ref{Disc} after presenting detailed experiments, results and implementation details in section \ref{EXP}. 

\section{Related Work}
\label{related}
One of the very first comparative studies on visible and thermal face recognition was performed by Socolinsky and Selinger \cite{socolinsky2002}. They concluded that "LWIR thermal imagery of human faces is not only a valid biometric,  but almost surely a superior one to comparable visible imagery." A good survey on single modal and cross-modal face recognition methods can be found in \cite{zhou2014recent}.

In the cross-modal (infrared-visible) face recognition scenario, most of the earlier efforts focus only on the NIR to visible matching \cite{yi2007, liao2009, Klare2010, lei2009}. Yi \etal \cite{yi2007} is one of very first to investigate NIR to visible face recognition. They used LDA and canonical correspondence analysis to perform linear regression between the NIR and visible images. Few methods have also focused on SWIR-to-visible face recognition \cite{ross2010, nicolo2012}. NIR or SWIR to visible face matching produces much better results as both the modalities are very similar because of the small spectral gap. Because of their limited use in night-time surveillance applications, further research focus is required in the thermal to visible matching domain.

Only recently, some interest in the thermal to visible face matching has emerged. In the thermal domain, most of these methods are evaluated in the MWIR to visible scenario. An interesting approach is by Li \etal to perform recognition by trying to synthesize an equivalent visible image from the thermal counterpart \cite{li2008}. Other interesting local feature based approaches include \cite{Klare13, bourlai2012, hu2014, hu2015}. Among these, Klare and Jain \cite{Klare13} proposed to represent the two modalities by storing only similar vectors as prototypes during training using a similarity kernel space termed as prototype random subspace. The most difficult acquisition scenario is the LWIR thermal images, because of the long wave range sensing, the produced images are usually of considerably lower resolution making it more challenging to match. Because of their higher emissivity sensing range they are also well suited to operate in total darkness. Attempts have been made to address the face matching in the LWIR to visible domain. Choi \etal \cite{choi2012} presented a partial least square `PLS' based regression framework to model the large modality difference in the PLS latent space. The most recent and best state-of-the-art results on the same dataset have been achieved by Hu \etal \cite{hu2015}. They used a discriminant PLS based approach by specifically building discriminant PLS gallery models for each subject by using thermal cross examples. They achieved a rank-1 identification of 50.4\% using two gallery images per subject and testing against all the probe LWIR thermal images. Finally, for facial feature detection in thermal images, Eslam \etal \cite{mostafa2013} presented a good method based on Haar features and Adaboost. They also performed  face recognition experiments, but used both the visible and thermal modalities in the gallery and probe set.

Deep and large neural networks have exhibited impressive results for visible face recognition. Most notably, Taigman \etal \cite{taigman2014} showed excellent results using convolutional neural network to learn the face representation. Hu \etal \cite{hu2014discriminative} employ a feed forward network for metric-learning using pairs of same and different visible images. In contrast, we use a  network to learn the non-linear regression projections between visible and thermal data without trying to learn any distance metric or image representation. A similar objective is employed in the domain adaptation based approaches e.g., \cite{ganin2014}, where the idea is to use the knowledge of one domain to aid the recognition in another image domain. Our model, while trying to learn a non-linear regression mapping, is simple and different from these in terms of the objective function and the way the network is used. We use a fully-connected feed forward network with the objective of directly learning a perceptual mapping between the two different image modalities. 

\section{Deep Perceptual Mapping (DPM)}
\label{DPM}
The large perceptual gap between thermal and visible images is both because of the modality and resolution difference. The relationship between the two modalities is highly non-linear and difficult to model. Previous approaches try to use non-linear function mapping e.g., using the kernel trick or directly estimating the manifold where the two could lie closer together. Since such a manifold learning depends on defining the projections based on a given function, they are highly data dependent and require a good approximation of the underlying distribution of the data. Such a requirement is hard to meet in practice, especially in the case of thermal-visible face recognition. Here, deep neural networks can learn, to some extent, the non-linear mapping by adjusting the projection coefficients in an iterative manner over the training set. As the projection coefficients are learned and not defined, it may be able to capture the data specific projections needed to bring the two modalities closer together. Based on this foresight we attempt to learn such projections by using a multilayer fully-connected feed forward neural network. 

\subsection{The DPM Model}
The goal of training the deep network is to learn the projections that can be used to bring the two modalities together. Typically, this would mean regressing the representation from one modality towards the other. Simple feed forward neural network provides a good fitting architecture for such an objective. We construct a deep network comprising $N+1$ layers with $m^{(k)}$ units in the $k$-th layer, where $k=1, 2, \cdots, N$. For an input of $x\in \mathbb{R}^{d}$, each layer will output a non-linear projection by using the learned projection matrix $\mathbf{W}$ and the non-linear activation function $g(\cdot)$. The output of the $k$-th hidden layer is $h^{(k)}=g(\mathbf{W}^{(k)}h^{(k-1)}+\mathbf{b}^{(k)})$, where $\mathbf{W}^{(k)}\in \mathbb{R}^{m^{(k)} \times m{(k-1)}}$ is the projection matrix to be learned in that layer, $\mathbf{b}^{(k)}\in \mathbb{R}^{m^{(k)}}$ is a bias vector and $g:\mathbb{R}^{m^{(k)}} \mapsto \mathbb{R}^{m^{(k)}}$ is the non-linear activation function. Note, $h^{(0)}$ is the input. Similarly, the output of the most top level hidden layer can be computed as:
\begin{equation}
\label{eq1}
\mathbf{H}(x)=h^{(N)}=g(\mathbf{W}^{(N)}h^{(N-1)}+\mathbf{b}^{(N)})
\end{equation}
where the mapping $\mathbf{H}:\mathbb{R}^{d} \mapsto \mathbb{R}^{m^{(N)}}$ is a parametric non-linear perceptual mapping function learned by the parameters $\mathbf{W}$ and $\mathbf{b}$ over all the network layers.
Since the objective of the model is to map one modality to the other, the final output layer of the network employs a linear mapping to the output in Equation \ref{eq1} to  obtain the final mapped output:
\begin{equation}
\label{eq2}
\bar{x}=s(\mathbf{W}^{(N+1)}\mathbf{H}(x))
\end{equation}
where $\mathbf{W}^{(N+1)} \in \mathbb{R}^{m^{(N)} \times d}$ and $s$ is the linear mapping. Equation \ref{eq2} provides the final perceptually mapped output $\bar{x}$ for an input $x$.
To determine the parameters $\mathbf{W}$ and $\mathbf{b}$ for such a mapping, our objective function must seek to minimize the perceptual difference between the visible and thermal training examples in the least mean square sense. We, therefore, formulate the DPM learning as the following optimization problem.
\begin{equation}
\label{eq3}
arg ~ \min_{W,b} ~~~ \mathbf{J}=\frac{1}{M}\sum_{i=1}^M (\bar{x}_i - t_i)^2 + \frac{\lambda}{N} \sum_{k=1}^N(\|\mathbf{W}^{(k)}\|_F^2 + \|\mathbf{b}^{(k)}\|_2^2) 
\end{equation}

The first term in the objective function corresponds to the simple squared loss between the network output $\bar{x}$ given the visible domain input and the corresponding training example $t$ from the thermal domain. The second term in the objective is the regularization term with $\lambda$ as the regularization parameter. $\|\mathbf{W}\|_F$ is the Frobenius norm of the projection matrix $\mathbf{W}$. Given a training set $\mathbf{X}=\left\{x_1,x_2,\cdots,x_M\right\}$ and $\mathbf{T}=\left\{t_1,t_2,\cdots,t_M\right\}$  from visible and thermal domains respectively, the objective of training is to minimize the function in equation \ref{eq3} with respect to the parameters $\mathbf{W}$ and $\mathbf{b}$.
 \\[0.5\baselineskip]\noindent\textbf{The DPM Training:} There are some important design considerations for a meaningful training of the DPM network to provide an effective mapping from visual to thermal domain. First, the model is sensitive to the number of input and output units. A high dimensional input would require a very high amount of training data to be able to reasonably map between the two modalities. We propose to use the densely computed feature representations from overlapping small regions in the images. This proves very effective, as not only the model is able to capture the differing local region's perceptual differences well but also alleviate the need of large training images and nicely present the input in relatively small dimensions. The training set $\mathbf{X}$ and $\mathbf{T}$, then, comprises of these vectors coming from the corresponding patches from the images of the same identity. Note, only using the corresponding images of the same identity ensures that the model will learn the only present differences due to the modality as the other appearance parameters \eg identity, expression, lighting etc., would most likely be the same. The network is trained, over the training set, by minimizing the loss in Equation \ref{eq3} \wrt the parameters. Figure \ref{fig:1} capsulizes this process. By computing the gradient of the loss $J$ in each iteration, the parameters are updated by standard back projection of error using stochastic gradient descent `SGD' method \cite{glorot2010}. For the non-linear activation function $g(z)$ in the hidden layers, we have used the hyperbolic tangent `$tanh$' as it worked better than other functions \eg sigmoid and ReLU on our problem. $tanh$ maps the input values between -1 to 1 using $g(z)= \frac{e^{2z}-1}{e^{2z}+1}$.

Before commencing training, the parameters are initialized according to a uniform random initialization. As suggested in \cite{glorot2010}, the bias $\mathbf{b}$ is initialized as $0$ and the $\mathbf{W}$ for each layer is initialized using the uniform distribution $\mathbf{W}^{(k)}\sim  U\left[-\frac{\sqrt6}{\sqrt{m^{(k)}+m^{(k-1)}}},\frac{\sqrt6}{\sqrt{m^{(k)}+m^{(k-1)}}}\right]$, where $m^{(k)}$ is the dimensionality (number of units) of that layer.   
\subsection{Thermal-Visible Face Matching}
After obtaining the mapping from visible to thermal domain, we can now pose the matching problem as that of comparing the thermal images with that of mapped visible data. Specifically, the mapped local descriptors from overlapping blocks of the visible gallery images are concatenated together to form a long vector. The resulting feature vector values are $L_2$-normalized and then matched with the similarly constructed vector directly from the probe thermal image. Note that, only visible gallery images are mapped through the learned projections using Equations \ref{eq1} and \ref{eq2} whereas the densely computed vectors from thermal images are directly concatenated into the final representation vector.
 
The presented set-up is ideal for the surveillance scenario as the gallery images can be processed and stored offline while at test time no transformation and overhead is necessary. As we will show in the next section, without any computational overhead the probes can be matched in real-time. Note, however, the  presented DPM is independent of the direction of mapping. We observed only slight performance variations using the opposite thermal to visible mapping.

The identity of the probe image is simply determined by computing the similarity with each of the stored gallery image vectors and assigning the identity for which the similarity is maximum. Since the vectors are $L_2$-normalized, the cosine similarity can simply be computed by the \textit{dot} product.   
\begin{equation}
d(\bar{x_i},t_j)=\bar{x_i}\cdot t_j \qquad\qquad \forall i=1,2,\cdots ,G
\end{equation}
where $t_j$ is the $j$-th constructed probe thermal image vector and $G$ is the total number of stored gallery vectors. Computationally, this is very efficient as at the test time, each incoming probe image can be compared with the whole gallery set $\mathbf{\bar{X}}$ by a single matrix-vector multiplication.

\section{Experimental Results}
\label{EXP}
\begin{figure}
    \includegraphics[width=\textwidth]{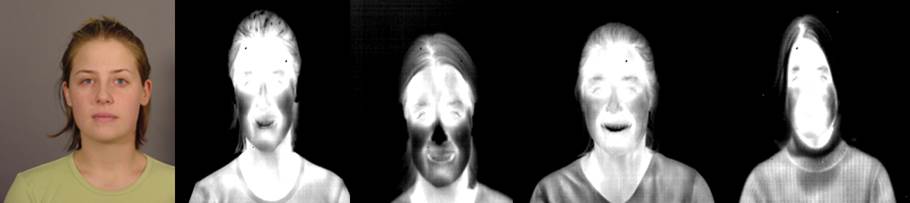}
    \caption{Problem: Matching thermal to the stored high resolution visible image. A wide modality gap exists between visible and thermal images of the same subject. Thermal images depict typical image variations, present, due to lighting, expressions and time-lapse.}
    \label{fig:2}
\end{figure}

In this section we present the evaluation of the proposed approach on a difficult LWIR-visible face recognition scenario. We report evaluations using typical identification and verification settings. The evaluation results, assessing the proposed DPM mapping, and the implementation details are discussed in the following subsections. 
\subsection{Database Description \& Implementation Details}
\textbf{UND Collection X1:} Only few publicly available datasets include thermal and corresponding visible facial acquisitions. Among these, University of Notre Dame's UND collection X1 is a challenging LWIR and visible facial corpus. The dataset was collected using a Merlin uncooled LWIR sensor and a high resolution visible color camera \cite{chen2005ir}. The resolution of the visible images is $1600\times1200$ pixels and the acquired LWIR thermal images is $312\times239$ pixels. This depicts the challenge not only due to the large modality gap but also due to a large resolution difference. The dataset includes images in three experimental settings: expression (neutral, smiling, laughing), lighting change (FERET lighting and mugshot) and time-lapse (subjects acquired in multiple sessions over time). Following the protocol used by the recently published competitive proposals on this dataset \cite{hu2015, choi2012}, we use all the available images in these three settings of the thermal probes to match against a single or multiple available high resolution visible gallery image/s. The dataset contains $4584$ images of 82 subjects distributed evenly in visible and thermal domain. To compare our results, we use the exact same training and testing partitioning as used in the previous methods. The pool of subjects is divided into two equal subsets: visible and thermal images from 41 subjects are used for the gallery and probe images (partition A), while the remaining 41 subjects were used to train the DPM model. Note that the training and test sets are disjoint both in terms of images and subject's identities. All thermal images from partition-A are used as probes to compute recognition performance. The images are aligned based on provided eye and mouth locations and facial regions are cropped to $110\times150$ pixels. Figure \ref{fig:2} shows some sample visible and corresponding thermal images of one subject in the database.
\\[0.5\baselineskip]\noindent \textbf{Implementation Details:} Here we provide the parameters used and related details for implementation. To compare the results to previous methods, we use a similar preprocessing for the thermal and visible images. We use the median filtering to remove the dead pixels in thermal images, then zero-mean normalization and afterwards used Difference of Gaussian `DoG' filtering to enhance edges and to further reduce the effect of lighting. Dense features are then pooled from overlapping blocks in the image. The recognition performance varies between 2-5\% based on the descriptor used and the block size/overlap. A large overlap will produce more redundant information causing the network to saturate early while the block size effects the amount of structural information the network can still preserve while trying to map only the perceptual/modality difference. Based on our extensive evaluations we provide here the parameters that work well in practice. We have used a block size of $20\times20$ with a stride of $8$ pixels at two-scales (Gaussian smoothed with $\sigma=0.6$ and $\sigma=1$) of the image. This results in $408$ descriptors per image. As many prior methods \eg \cite{Klare13, hu2014, hu2015, choi2012} overwhelmingly stated the better performance of both the SIFT and HOG features in the thermal-visible face recognition. We experimented with both SIFT and HOG feature descriptors. However, here we present results with SIFT features as HOG produces $3\%$ inferior results than SIFT.

For the DPM model training, the number of units in the input/output and the hidden layers is important and varies the result. We use PCA to decorrelate and reduce the features to 64 dimensions. Each descriptor is further embedded with the block center position $(x,y)$, measured from the image center, to have the spatial position intact, this helps the model to better learn the local region specific transformations in the hidden layers. The number of units in the input/output layers are, therefore, $66$.

For the DPM network, we obtained better results with a minimum of 3-layers ($N=3$ in Equations \ref{eq1} and \ref{eq2}). Although, for comparison, we also report performance with using $N=2$ \ie one hidden layer. As our results show, the DPM mapping is highly effective even with this shallow DPM configuration. The minimum number of units in the hidden layers,  to ensure good results, is set empirically by using different combinations. Here, we report results using 200 units in each of the two hidden layers for the deep configuration. While in the case of shallow (1 hidden layer configuration), $1000$ hidden units in the single layer provide close results. It is worth noticing, that the the results vary slightly about 1-4\% using different number of hidden units in the range $200-1000$ in a layer. Finally the DPM is trained by pooling over 1 million patch vectors from the training set (1396 visible and 1396 thermal images).
\subsection{Results}
\begin{figure}
  \begin{minipage}[t]{0.48\textwidth}  
    \includegraphics[width=\textwidth]{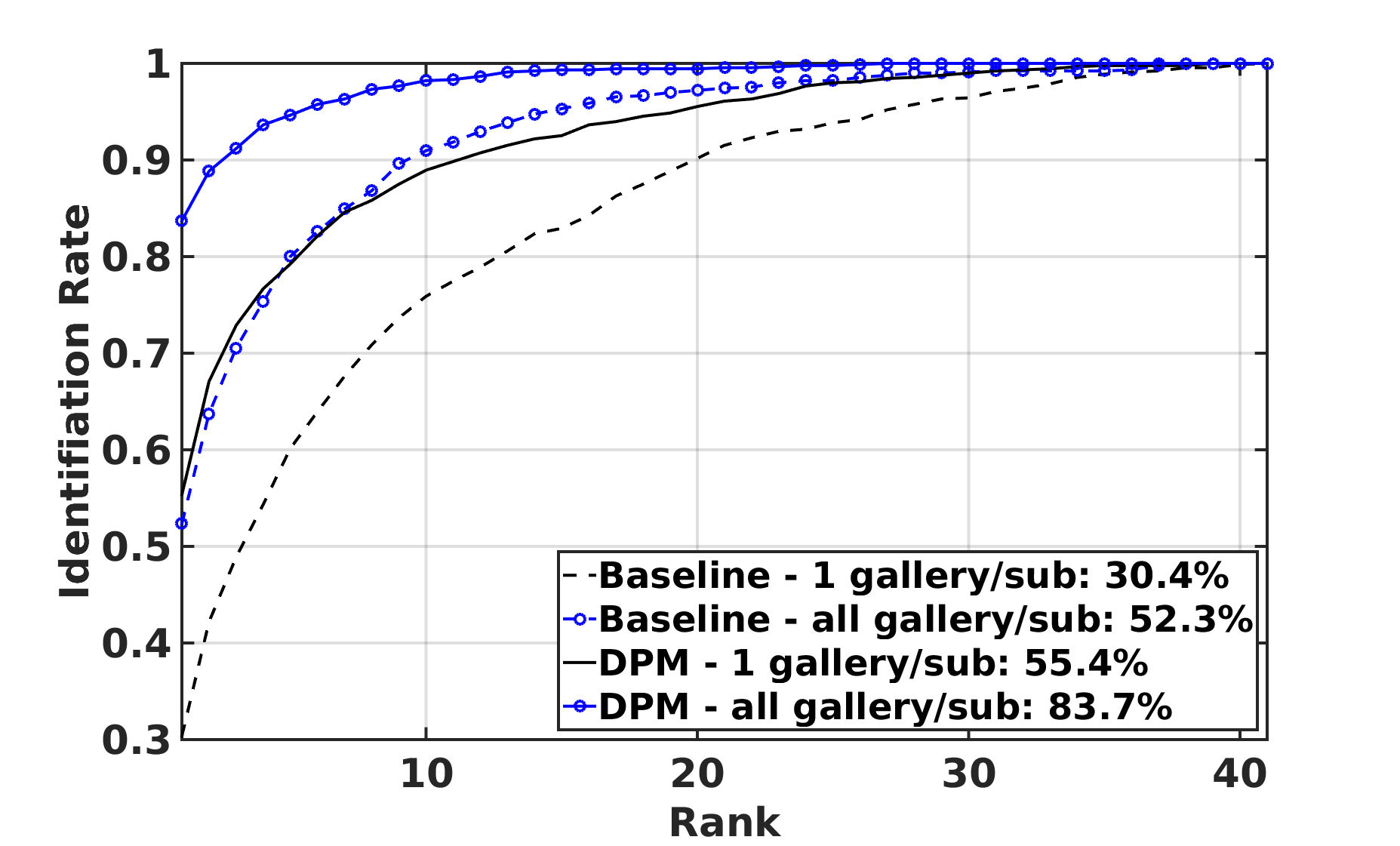}
	\caption{Rank-wise score: comparison of baseline and DPM performance with 1-visible image/sub and all available visible images/sub in the gallery.}
    \label{fig:3}
  \end{minipage}
  \hfill
  \begin{minipage}[t]{0.48\textwidth}
    \includegraphics[width=\textwidth]{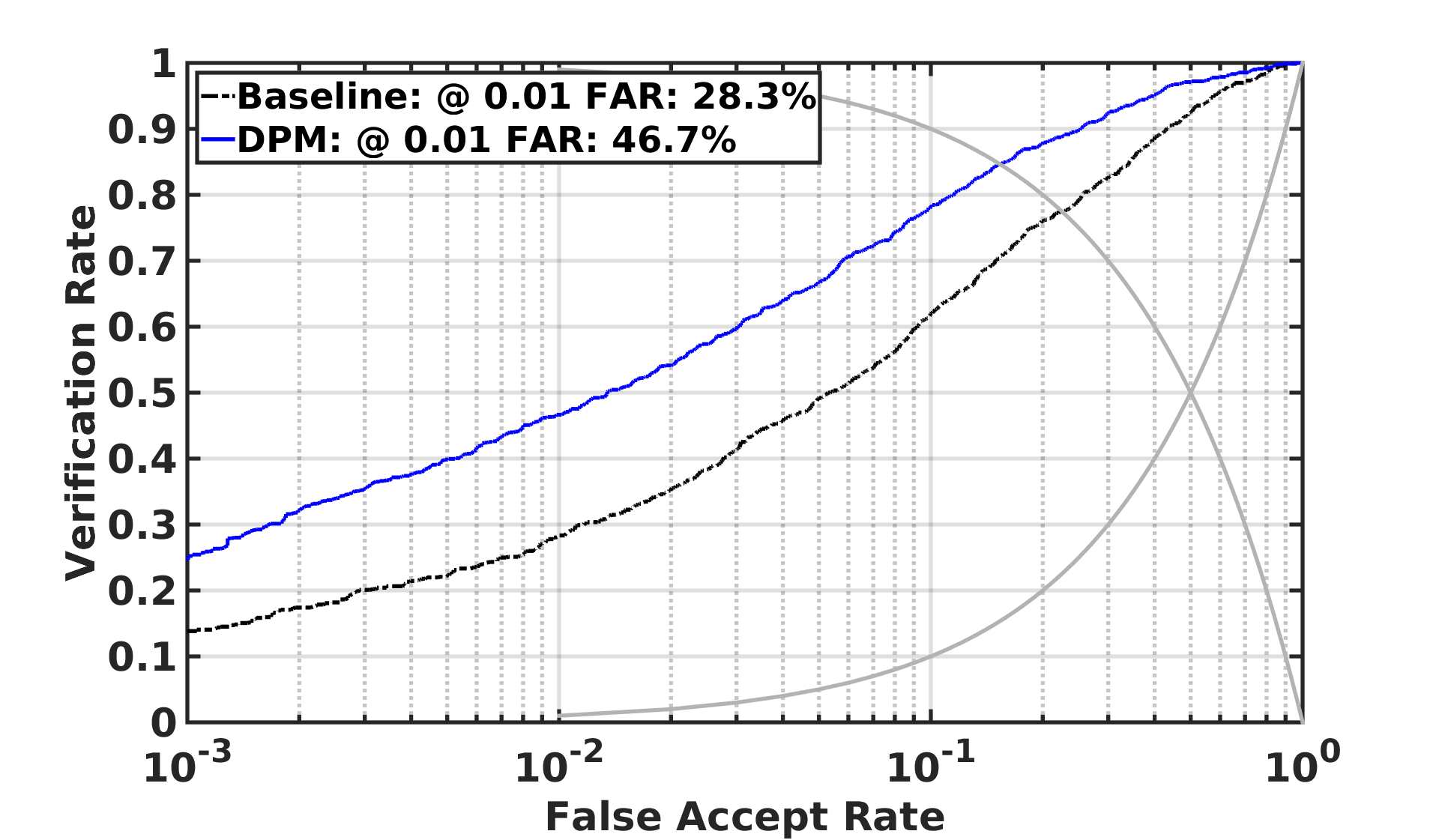}
    \caption{Verification performance thermal-visible: all thermal probes with 2 visible image/sub in gallery. ROC curves from 1792 genuine and 71680 imposter attempts.}
    \label{fig:4}
  \end{minipage}
\end{figure}

\textbf{Baseline:} We provide results of our evaluations using the settings described before. As baseline we use the same concatenated SIFT features but without the DPM mapping. This would directly enable us to compare and appreciate the power of the proposed model.
\\[0.5\baselineskip]\noindent\textbf{Identification Evaluation:} Table \ref{table:1} presents the comparison results (Rank-1 identification score) with the baseline and the best published state-of-the-art results on the dataset. We present results using three different gallery settings. The most difficult and restricted setting with only one neutral visible image per subject in the gallery, using two visible images/subject in the gallery ( the first two images, neutral and with smiling expression) and using all the available visible images/subject as gallery. Note that all the available thermal images/subject are used as probes. The number of images for each subject in the thermal probe set varies from a minimum of $4$ to a maximum of $40$ depicting appearance variations due to expressions, lighting and time-lapse. As the results show, we improve the state-of-the-art best published results of \cite{hu2015} by more than 10\% in all cases. We also report the results using shallow DPM configuration (with 1-layer). Our results depicts that even this shallow DPM configuration well surpasses the best published results on this dataset.
Figure \ref{fig:3} presents the cumulative match characteristics `CMC' curves to measure the rank-wise identification performance of the presented method. It shows the effectiveness of the proposed DPM mapping in comparison with the baseline (same features without the DPM mapping).
\\[0.5\baselineskip]\noindent\textbf{Verification Evaluation:} We also evaluate the verification accuracy of such cross-modal scenario. We report here the results by using 2 visible images/subject in the gallery. Given the size of our thermal probe set, this amounts to having $1792$ genuine and $71680$ imposter attempts. Figure \ref{fig:4} presents the ROC curve measuring the verification performance gain using the DPM mapping.
\begin{table}[t]
\centering
\begin{tabular}{@{}lccc@{}}
\toprule
                                 & \multicolumn{3}{c}{Gallery size: \# of visible images/subject} \\ \cmidrule(l){2-4} 
                                 & 1/subject         & 2/subject        & all available/subject   \\ \midrule
Baseline (same features w/o DPM) & 30.36             & 35.60            & 52.34                   \\
Choi \etal \cite{choi2012} & -                 & 49.9             & -                       \\
Hu \etal \cite{hu2015} & 46.3              & 50.4             & 72.7                    \\
DPM-\textit{1 hidden-layer} (ours)             & 50.67              & 58.26             & 79.57                   
\\
\textbf{DPM  (Ours)}             & \textbf{55.36}    & \textbf{60.83}   & \textbf{83.73}          \\ \bottomrule
\end{tabular}
\caption{Comparison of Rank-1 Identification accuracy (\%) using all thermal images as probes and visible images in the gallery.}
\label{table:1}
\end{table}
\\[0.5\baselineskip]\noindent\textbf{Effect of modality gap:} Finally, we present the experiment to measure the effect of modality gap. Keeping everything fixed \ie using the same baseline features and settings, we compute the rank-1 identification score within the same modality. We use one-thermal image/subject (the neutral frontal image) to form the thermal gallery and tested against all the remaining thermal images as probes. This is the same setting as we have used in the cross-modal thermal-visible case. We obtain $89.7\%$ rank-1 score in the Thermal-Thermal identification scenario. While as, also, shown earlier in Table \ref{table:1} the rank-1 identification in the corresponding Thermal-visible scenario (using the same baseline features) is $30.3\%$. This amounts to the performance drop, purely due to modality change, of about $59\%$. This reflects the challenging nature of the problem and the existing research gap to tackle this. Table \ref{table:2} compares the results of Thermal-Thermal and Thermal-visible identification to quantify the effect of modality gap. As shown , with DPM on the same features, the performance is improved by $25\%$. This amounts to bridging the existing modality gap of $59\%$ by more than $40\%$.
\\[0.5\baselineskip]\noindent\textbf{Computational Time:} Training the DPM on 12 cores 3.2-GHz CPU takes between $1-1.5$ hours on MATLAB. Preprocessing, features extraction and mapping using DPM only takes $45 ms$ for one image. This is even less in the testing case since no mapping is required for thermal images. At test time identifying one probe only takes $35 ms$. Since we are using just the dot product between the extracted probe vector and the gallery set, this is therefore, very fast and capable of running in real-time at $\sim 28$ fps.

\begin{table}[t]
\centering
\begin{tabular}{@{}cccc@{}}
\toprule
\multicolumn{4}{c}{Effect of Modality gap: Performance with 1 Gallery image/subject} \\ \midrule
Thermal-Thermal   & Thermal-Visible  & Thermal-Visible (via DPM)  & Modality-gap bridged  \\
89.47             & 30.36            & 55.36                       & $\sim 42 \%$               \\ \bottomrule
\end{tabular}
\caption{Performance drop due to Modality gap: Rank-1 identification using 1 image/subject as gallery in Thermal-Thermal and Thermal-Visible matching using baseline features.}
\label{table:2}
\end{table}

\section{Discussion \& Conclusions}
\label{Disc}
While providing quantitative evidence of the effectiveness of the proposed approach, we can also compare our results qualitatively with some other methods that evaluate the thermal-visible face recognition on similar private datasets. Bourlei \etal \citep{bourlai2012} reported results on a MWIR-visible face recognition task using a $39$ subject dataset with four visible images per subject and three MWIR images per subject as probes with $1024\times1024$ pixel resolution. They reported a rank-1 performance of $53.9\%$. Klare and Jain \cite{Klare13} evaluated on a private dataset using MWIR images of 333 subjects (total number of images and other condition not known) and an extended large visible gallery. The rank-1 recognition of 49.2\% is reported. Note that MWIR-visible has relatively less pronounced modality gap and much higher resolution as that of LWIR-visible. One can qualitatively see the power of the presented DPM approach on a much harder problem.  

Thermal-visible face recognition is a very difficult problem due to the inherent large modality difference. Our presented method is very effective and has benefits for many related applied computer vision and domain adaptation problems from image matching, detection to recognition. Similarly it is also very attractive for remote sensing applications, where it is a common problem to register and match images from different modalities (\eg coming from different satellites). This can be used as a very effective prior step to bridge the modality and/or large resolution difference before using the representation vectors in any common learning scheme. Conclusively, the presented DPM approach is very useful, easy to train, and real-time capable due to very less computational overhead and provides a practical solution for a large surveillance and military application industry. 

\section*{Acknowledgements}
This work was partially funded by the German Federal Ministry of Education and Research (BMBF) under `IntenC' cooperation program contract no. 01DL13016. The views expressed herein do not necessarily reflect those of BMBF.

\bibliography{bibours}

\end{document}